\documentclass[10pt, conference, compsocconf]{IEEEtran} 
\pdfoutput=1

\usepackage{cite}
\usepackage[pdftex]{graphicx}
\usepackage{amsmath,amsfonts}
\usepackage{bm}

\newcommand{\RR}{\mathbb{R}}
\providecommand{\B}[1]{\bm{#1}}
\def \lb {{\langle}}
\def \rb {{\rangle}}

\usepackage{url}

\newcommand{\sgn}{\operatorname{sgn}}

\begin{document}
%
\title{Improved brain pattern recovery through ranking approaches}



\author{

\IEEEauthorblockN{Fabian Pedregosa \IEEEauthorrefmark{1}\IEEEauthorrefmark{2}\IEEEauthorrefmark{4},
Elodie Cauvet \IEEEauthorrefmark{3}\IEEEauthorrefmark{2},
Ga\"el Varoquaux\IEEEauthorrefmark{1}\IEEEauthorrefmark{2},
Christophe Pallier \IEEEauthorrefmark{3}\IEEEauthorrefmark{1}\IEEEauthorrefmark{2},
Bertrand Thirion\IEEEauthorrefmark{1}\IEEEauthorrefmark{2}}  and
Alexandre Gramfort\IEEEauthorrefmark{1}\IEEEauthorrefmark{2},
\\
\IEEEauthorblockA{
\IEEEauthorrefmark{1}Parietal Team, INRIA Saclay-\^{I}le-de-France, Saclay, France\\}
\IEEEauthorblockA{
\IEEEauthorrefmark{2}
CEA, DSV, I\textsuperscript{2}BM, Neurospin b\^{a}t 145,
91191 Gif-Sur-Yvette, France
\\}
\IEEEauthorblockA{\IEEEauthorrefmark{3}
Inserm, U992, Neurospin b\^{a}t 145, 91191 Gif-Sur-Yvette, France\\}
\IEEEauthorblockA{\IEEEauthorrefmark{4} 
SIERRA Team, INRIA Paris - Rocquencourt, Paris, France
}}

\maketitle

\begin{abstract}
  Inferring the functional specificity of brain regions from functional Magnetic Resonance  Images (fMRI) data is a
  challenging statistical problem. While the General Linear Model
  (GLM) remains the standard approach for brain mapping, supervised
  learning techniques (\emph{a.k.a.}  decoding) have proven to be
  useful to capture multivariate statistical effects distributed
  across voxels and brain regions. Up to now, much effort has been
  made to improve decoding by incorporating prior knowledge in the
  form of a particular regularization term. In this paper we
  demonstrate that further improvement can be made by accounting for
  non-linearities using a ranking approach rather than the commonly used 
  least-square regression. Through simulation, we compare
  the recovery properties of our approach to
  linear models commonly used in fMRI based decoding. We demonstrate the
  superiority of ranking with a real fMRI dataset.
\end{abstract}


%
\IEEEpeerreviewmaketitle

\section{Introduction}
The prediction of behavioral information or cognitive states from
brain activation images such as those obtained with fMRI can be used
to assess the specificity of several brain regions for certain
cognitive or perceptual functions. This kind of analysis is
implemented by learning a classifier or regression function that fits a
given {\it target} variable given fMRI activations. The accuracy of
this prediction depends on whether it uses the relevant variables
{\it i.e.} the correct brain regions. {\it Recovering} the truly
predictive pattern has proven to be challenging from a statistical
point of view: the high dimensionality of the data together with the
limited number of images makes the problem of brain pattern recovery
an ill-posed problem.

So far, the approaches proposed to address this issue have relied on
linear models, with univariate, \emph{i.e.}
voxel-based, Anova (analysis of variance) for hypothesis testing, or,
for predictive modeling, with 
the choice of a regularizer using a priori domain-specific knowledge,
such as the $\ell_1$-norm
to promote sparsity
\cite{yamashita2008,Ryali_Supekar_Abrams_Menon_2010}, total
variation to promote spatial smoothness \cite{Michel2011}.
Various data fit terms have been used, Logistic Regression (LR)
\cite{Ryali_Supekar_Abrams_Menon_2010}, Linear SVM \cite{mouraomiranda2005}, 
Lasso \cite{yamashita2008}. While Linear SVM and LR
cannot address the recovery problem for multiclass problems, linear
regression models
assume a linear relationship between the quantity to predict and the amplitude
of the fMRI signals.
If the linear relationship does not hold in practice,
then the estimation of the predictive patterns
may suffer from a loss of statistical power.
This can be particularly relevant with Blood Oxygen-Level Dependent
(BOLD) signals observed in fMRI, where a saturation effect is expected
as the level of signal increases.

When targets to predict consist of ordered values, as in a parametric
design, such as
clinical scores, pain levels or the complexity of a
cognitive task, the response to these different conditions can reflect
the non-linearities in the data.
In such situation, we propose to
use a data fit term, known as loss function, not relying on an assumption
of linearity but only of increasing response. We show on simulations 
that this new
formulation opens the door to capturing the non-linearity and leads to
better recovery of the predictive brain patterns. On an fMRI dataset
we show that
the new formulation leads to models with better recovery properties.



\paragraph{Notations}

We write vectors in bold, $\B{a} \in \RR^{n}$, matrices with capital
bold letters, $\B{A} \in \RR^{n\times p}$. The dot product between
two vectors is denoted $\lb \B{a}, \B{b} \rb$. We denote by
$\|\mathbf{a}\|  = \sqrt{\lb \B{a}, \B{a} \rb}$ the $\ell_2$ norm of a
vector.

\section{Learning a linear model from fMRI data}

Following standard statistical learning notations we denote by
$\B{x}_i\in\RR^p$, $1 \leq i \leq n$, the data and $y_i \in
\mathcal{Y}$ the target variables.  In this paper, we aim at learning
a weight vector $\mathbf{w} \in {\RR}^p$ such that the prediction of
$\B{y}$ can be non-linearly related to the value of $\mathbf{w}^T
\mathbf{x}$. The vector $\mathbf{w}$ corresponds here to a brain map that
can be represented in brain space as a volume for visualization of the
predictive pattern of voxels. It is useful to rewrite these quantities
in matrix form; more precisely, we denote by $\mathbf{X} \in \RR^{n
  \times p}$ the design matrix assembled from $n$ fMRI volumes and by
$\B{y} \in \RR^n$ the corresponding $n$ targets. In other words, each
row of $\mathbf{X}$ is a $p$-dimensional sample, i.e., an activation
map of $p$ voxels related to one stimulus presentation.

A standard approach to perform the estimation of $\mathbf{w}$
leads to the following minimization problem
\begin{equation}
\bold{\hat{w}} = \mathrm{arg}\!\min_{\bold{w}}\;
\mathcal{L}(\B{y}, \B{X}, \B{w}) + \lambda \,\Omega(\bold{w}) \;\;,\;\; \lambda \geq 0
\label{eq:opt_pb}
\end{equation}
where $\lambda \,\Omega(\bold{w})$ is the regularization term and
$\mathcal{L}(\bold{y},\bold{X}, \bold{w})$ is
the loss function. The parameter $\lambda$
balances the loss function and the penalty $\Omega(\bold{w})$.

If the explained variable is a linear combination of the images,
$\mathbf{y} = \mathbf{X} \mathbf{w} + \mathbf{\B{\epsilon}}$, we
can estimate $\hat{\mathbf{w}}$ using the mean squared error loss
function $\mathcal{L}(\bold{y},\bold{X}, \bold{w}) = \| \B{y} -
\mathbf{X} \mathbf{w}\|^2$. However, with fMRI the linearity
assumption may not be valid. Instead, we model our explained variable
as $\mathbf{y} = F(\mathbf{X} \mathbf{w}) + \mathbf{\B{\epsilon}}$,
where $F$ is a non-decreasing function.



We introduce the use of pairwise loss functions. These loss functions 
only assume the target values to be a non-decreasing
function of the data. They have been widely used in
ranking, a type of supervised machine learning problem whose goal is
to automatically construct an order from the training data.
A pairwise loss function operates on pairs of
images: given a pair of images $(\mathbf{x}_i, \mathbf{y}_i)$ and
$(\mathbf{x}_j, \mathbf{y}_j)$, $\mathbf{y}_i \neq \mathbf{y}_j$ we
build a model that predicts the sign of $\mathbf{y}_i -
\mathbf{y}_j$.

Let $\mathcal{I}$ denote the index set of all considered pairs. Note
that in some settings it might be important to restrict ourselves to a
selected subgroup of all pairs, {\it e.g.} to the pairs of images of a
single subject or to the pairs of images corresponding to a single
session. For this purpose we define $a_{ij} \in \RR, (i, j) \in
\mathcal{I}$ to be a weight associated to each pair.  We will now
present the pairwise loss functions used in this article:
\begin {itemize}

\item {\it Pairwise hinge loss} \cite{Herbrich2000}. This 
  is a natural extension of the loss function used by Support
  Vector Machines and has been successfully used in information
  retrieval \cite{Cao:2006:ARS:1148170.1148205}.
\begin{equation}
\sum\limits_{(i, j) \in \mathcal{I}}^{} a_{ij}[1 - \mathbf{w}^T(\mathbf{x}_i - \mathbf{x}_j)]_{+}
\label{eq:pairwise_hinge}
\end{equation}
where $[z]_{+} = \max\{z, 0\}$.

\item {\it Pairwise logistic loss}
  \cite{Dekel_Manning_Singer_2003}. This is the pairwise extension of
  the logistic regression loss function.
\begin{equation}
\sum\limits_{(i, j) \in \mathcal{I}}^{} a_{ij}\log(1 + \exp(\mathbf{w}^T(\mathbf{x}_i - \mathbf{x}_j)))
\label{eq:pairwise_logistic}
\end{equation}



\end {itemize}

When noise is present in the model, the order of two samples might get
inverted, a phenomenon known as {\it label switching}. Because this only
affects labels that lie close, it is natural to penalize
more the misclassification of distant labels. By setting the sample
weights to $a_{ij}$ to a value that increases as labels become more
separated, we become more robust to label switching. In the case of
hinge loss functions, several strategies for choosing the appropriate
weights are discussed in \cite{Cao:2006:ARS:1148170.1148205}.

On the implementation side, both pairwise hinge loss and pairwise
logistic loss can be implemented on top of existing SVM and Logistic
Regression solvers, respectively, by taking the difference of pairs as
input values. In practice, we used the liblinear
\cite{Fan_Chang_Hsieh_Wang_Lin_2008} library via the scikit-learn
\cite{scikit-learn} library.


\section{Simulation}

\begin{figure}[!t]
\centering
\includegraphics[width=2.5in]{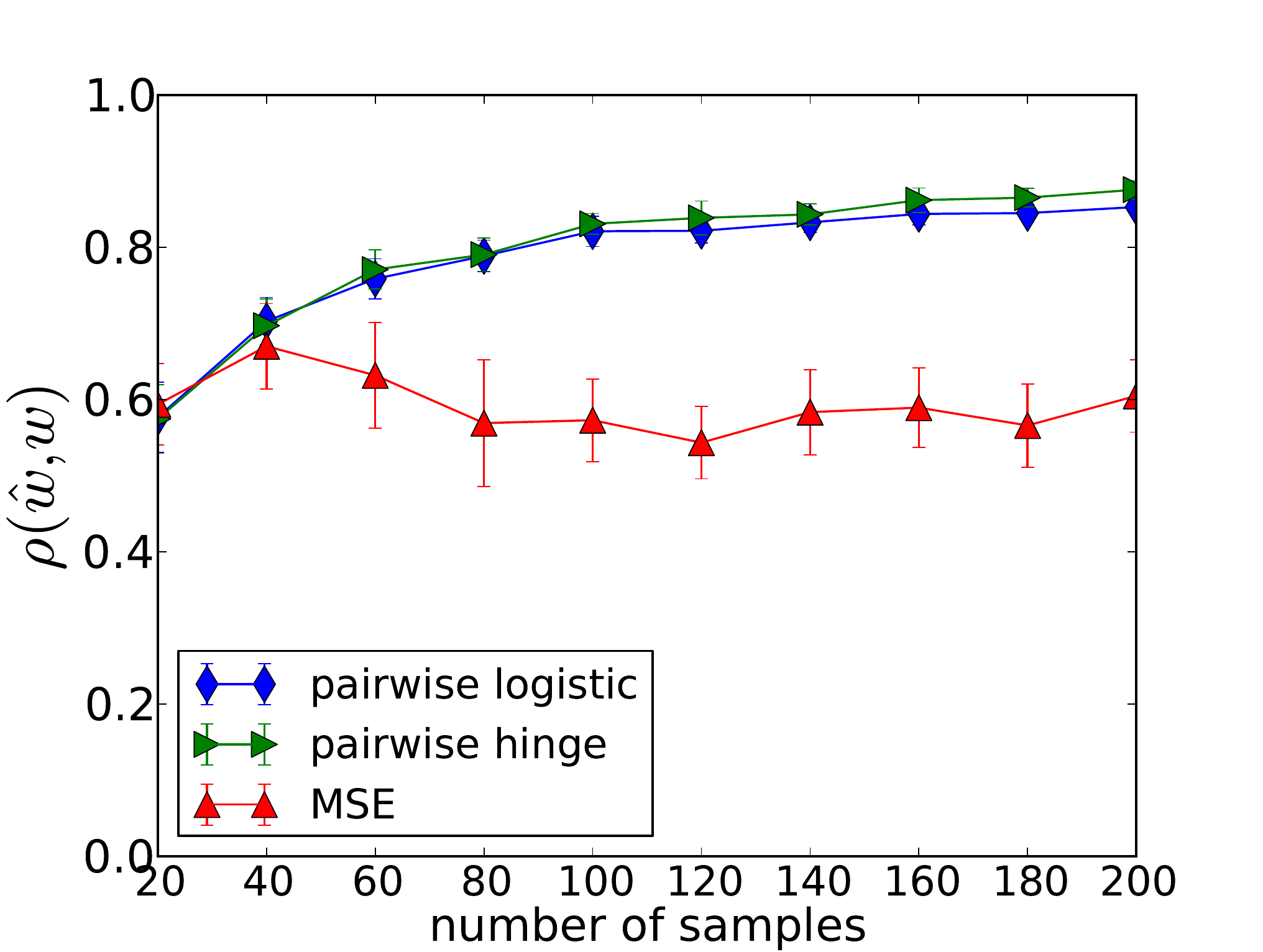}
\includegraphics[width=2.5in]{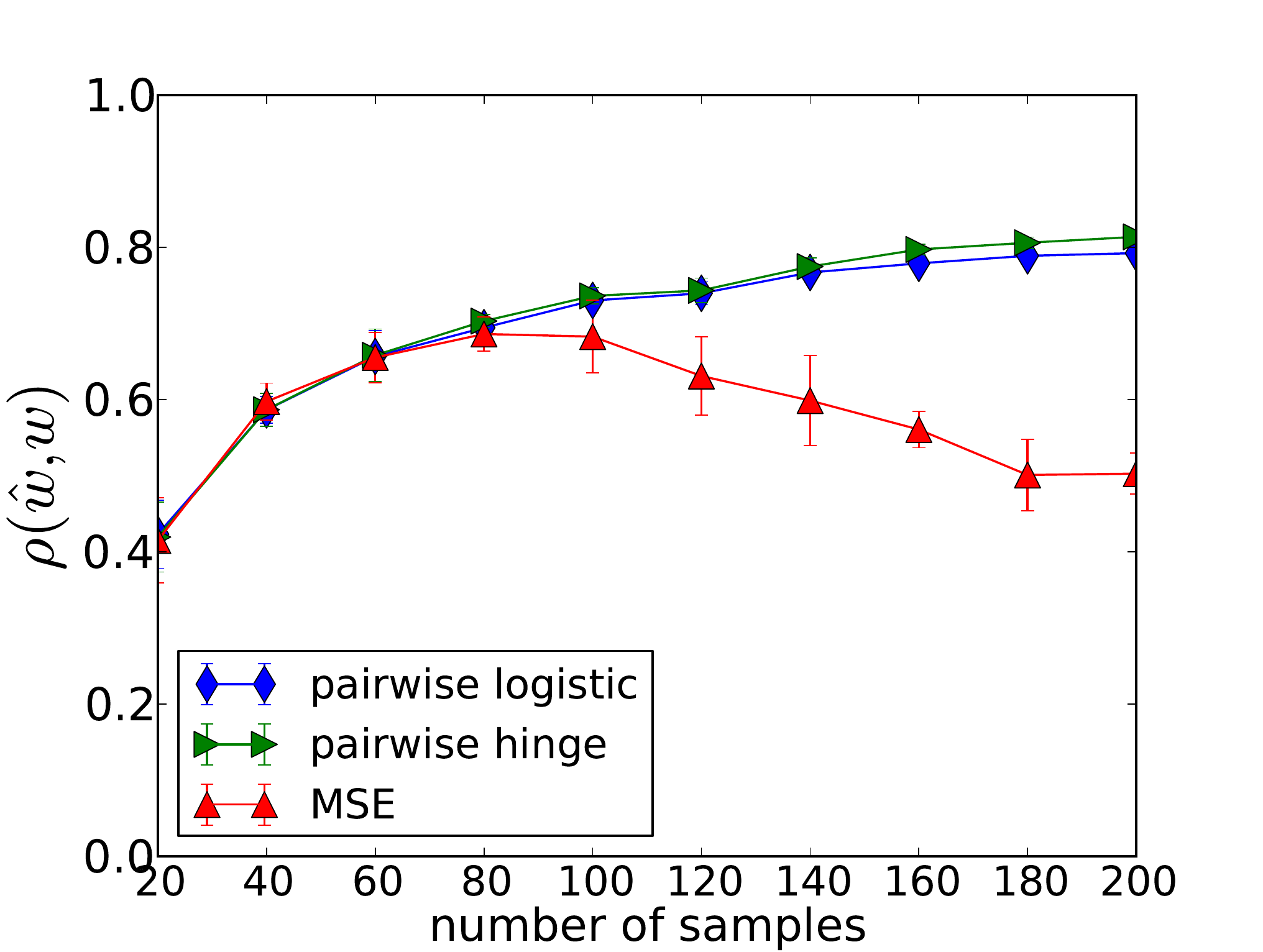}
\caption{Correlation between the estimated vector $\hat{\B{w}}$ and the
  ground truth $\B{w}$
  for different loss functions as the number of considered samples
  increases (higher is better) for dimensions $5 \times 5 \times 5$
  and $7 \times 7 \times 7$ respectively. Pairwise loss functions
  outperform linear regression as the number of samples increases and
  tend to a perfect recovery.}
\label{fig:rank_loss}
\end{figure}

\begin{figure}[!t]
\centering
\includegraphics[width=2.5in]{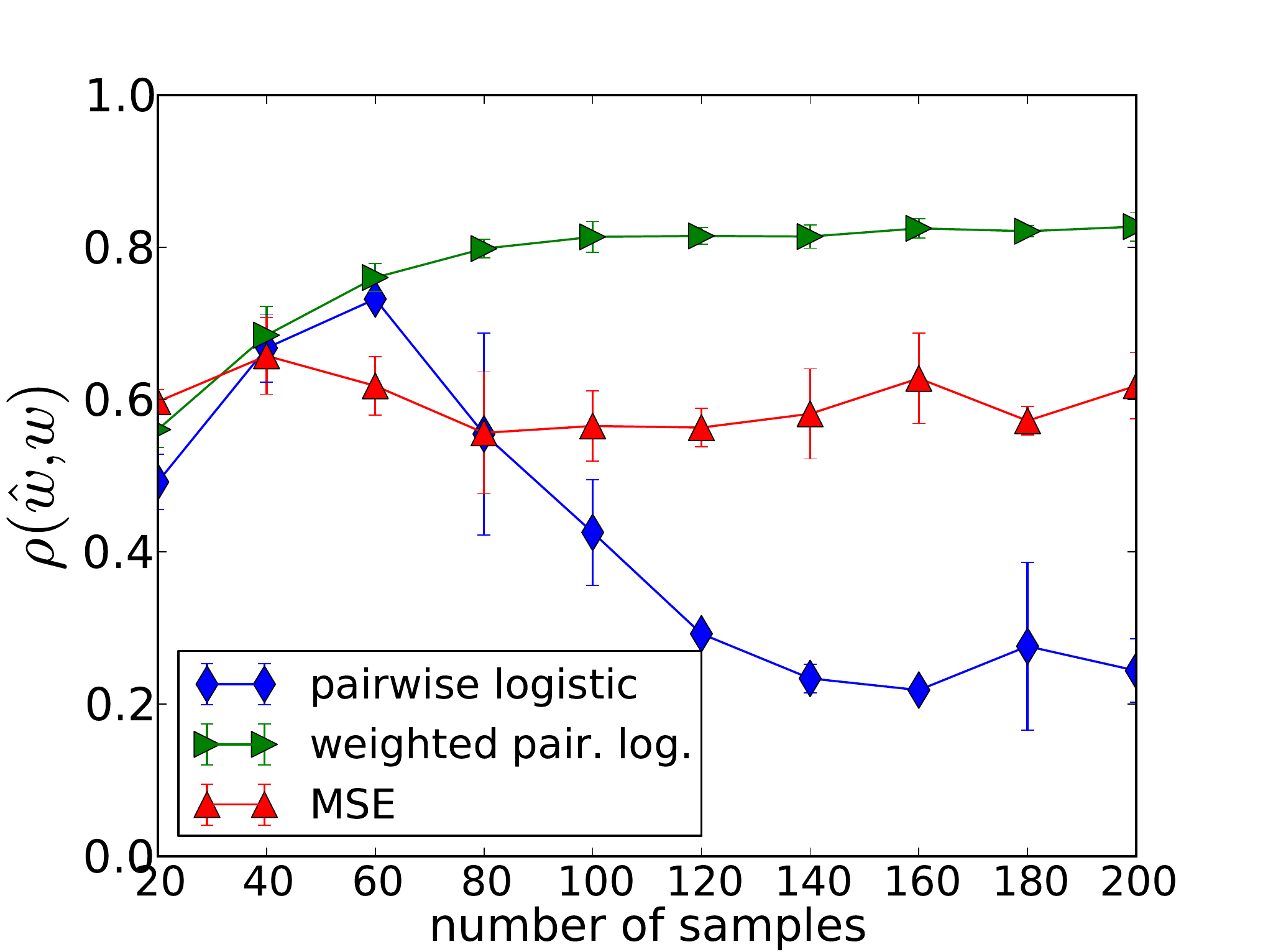}
\caption{Correlation between the estimated vector $\hat{\B{w}}$ and the
  ground truth $\B{w}$
  for different loss functions with a noise level of 5\%. Appropriately
  setting the weights plays a major role in robustness. Without the
  correct weighting and under noisy conditions, pairwise logistic loss
  function fails to recover the correct model. }
\label{fig:rank_loss2}
\end{figure}

\paragraph{Data generation} The simulated data $\mathbf{X}$ contains
volumes of size $5 \times 5 \times 5$ and $7 \times 7 \times 7$, each
one consisting of Gaussian white noise smoothed by a Gaussian kernel
with standard deviation of 2 voxels. This mimics the spatial
correlation structure observed in real fMRI data. The simulated vector
of coefficients $\mathbf{w}$ has a support restricted to four cubic
Regions of Interest (ROIs) of size ($2 \times 2 \times 2$). The values
of $\mathbf{w}$ restricted to these ROIs are $\{1, 1, -1, -1\}$.

The target variable $\mathbf{y}_{l} \in \RR^n$ is simulated as a linear
model:
\begin{eqnarray}
    \mathbf{y}_{l} = \mathbf{X}\mathbf{w} + \B{\epsilon}
\label{linear-model-with-noise}
\end{eqnarray}
where the noise $\B{\epsilon}_i \in [-\frac{\sigma}{2}, \frac{\sigma}{2}]$
follows a uniform distribution. $\sigma$ is chosen such that the
signal-to-noise ratio verifies $\|\B{\epsilon}\|/\|\B{X}\B{w}\| =
5\%$. We then define another target variable $\mathbf{y}_{nl}$ to be
a sigmoid function of $\mathbf{y}_{l}$, that is,
\begin{eqnarray}
\label{sigmoid}
    \mathbf{y}_{nl} = \frac{1}{1 + \exp{(-\mathbf{y}_{l}})}
\end{eqnarray}

For each of the loss functions introduced earlier and mean squared
error, we compute the correlation coefficient $\rho( \B{w},
\hat{\B{w}}) = \lb \B{w}, \hat{\B{w}}\rb / (\|\B{w}\| \|\hat{\B{w}}\|)$.
This gives us the goodness of fit for the estimated
$\hat{\B{w}}$. A method with correlation coefficient of 1 is able to
recover perfectly the ground truth. Since we are interested in the
estimation of $\B{w}$, we restricted ourselves to linear
models, leaving out models such as kernel ridge or support vector
regression with Gaussian kernel.

For all models we added $\ell_2$ regularization in the form of
$\lambda \|\B{w}\|^2$ and we cross-validated $\lambda$ separately for
each loss. In this setting, the model with mean squared error loss is
equivalent to {\it ridge regression}.  In order to focus on the effect
of non-linearity, we first considered a noiseless simulation using
trivial weights $a_{ij} = 1$.

Once we learned a vector $\hat{\B{w}}$ for each method we compute the
correlation coefficient with the ground truth for different sizes of the
training data. The experiment is repeated 10 times with different
initialization of the pseudorandom number generator. We compute
errorbars and show the results in figure~\ref{fig:rank_loss}. As the
number of samples increases, the linear model stalls and pairwise loss
functions outperform MSE on both $5 \times 5 \times 5$ and $7 \times 7
\times 7$ dimension. As expected, the higher dimensionality of
the second simulation makes the correlation coefficient decrease.
However, unlike MSE, ranking tends to a
perfect recovery as the number of samples increases.
Both pairwise loss functions perform equivalently
and have a significatively higher correlation coefficient than MSE. In
the rest of the paper we will use pairwise logistic as loss
function. As a result, pairwise loss functions should be preferred
over MSE in situations where underlying model is non-linear. Notice
that we fixed the non-linearity to be a sigmoid function, but the
pairwise loss functions only assume that this function is non-decreasing.
Unlike linear regression models, pairwise loss functions are indeed able
to learn the structure on the non-linear transform $F$.

We now consider the model with noise as in
\eqref{linear-model-with-noise} and use non-trivial weights $a_{ij}$. To
account for label switching, we set $a_{ij}$ to zero for pairs with
too similar labels:
\begin{eqnarray} \label{eq:loss}
a_{ij} = \begin{cases}
0 & \textnormal{ if } |\B{y}_i - \B{y}_j| < \sigma \\
1  & \textnormal{ otherwise} \;.
\end{cases}
\end{eqnarray}
In the case of discrete values, this would be equivalent to zeroing
weights for which the labels are adjacent. We now compute the
correlation coefficient for weighted and unweighted pairwise logistic
models and linear ridge regression model. The result can be seen in
figure \ref{fig:rank_loss2} for dimension $5 \times 5 \times 5$. The
unweighted logistic model breaks down in presence of noise and
performs worse than linear ridge regression. On the other hand,
appropriately setting the weights $a_{ij}$ has a major effect on
robustness, where this model outperforms MSE in a noisy setting. Note
also that weighted pairwise logistic has smaller variance than MSE.

\section{Results on fMRI data}

This dataset, described in~\cite{Cauvet2012}, consists of
34 healthy volunteers scanned while listening to 16 words sentences
with five different levels of complexity. These were 1 word
constituent phrases (the simplest), 2 words, 4 words, 8 words and 16
words respectively, corresponding to 5 levels of complexity which was
used as class label in our experiments. To clarify, a sentence with 16
words using 2 words constituents is formed by a series of 8 pairs of
words. Words in each pair have a common meaning but there is no meaning
between each pair. A sentence has therefore the highest complexity
when all the 16 words form a meaningful sentence.

The dataset consists of 8 manually labeled ROIs, some informative and
some not. For each ROI separately, we split the data into 60\%
training samples, 20\% for parameter selection and 20\% for
validation. We trained a pairwise logistic model and set the $\ell_2$
regularization by cross validation. We choose $a_{ij}$ to be zero if
classes are adjacent, i.e. if $|\B{y}_i - \B{y}_j| \leq 1$ and if
$\B{x}_i$ and $\B{x}_j$ do not belong to the same subject, in order to
consider exclusively non-adjacent pairs of images from the same
subject. In all other cases, $a_{ij}$ was set to one. We computed the
generalization score on the validation set as the mean number of
inversions with respect to the order in labels, i.e. the sign flips
$\sgn((\B{X}_i - \B{X}_j)\hat{\B{w}}) \neq \sgn(\B{y}_i - \B{y}_j)$
for all pairs of images in the validation set.

\begin{figure}[!t]
\centering
\includegraphics[width=2.5in]{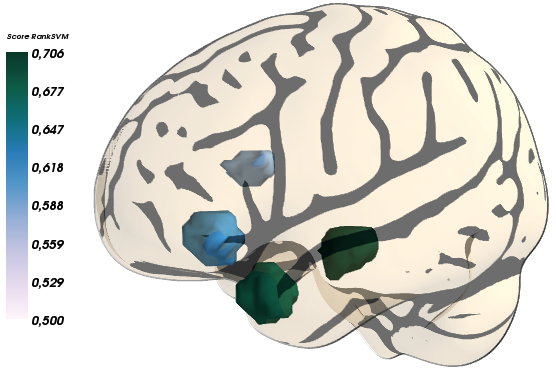}
        \caption{Scores obtained with the pairwise logistic on the 4 different ROIs.
        The regions with the best predictive power are the temporal pole
        the anterior superior temporal sulcus.}
\label{fig:scores_brain}
\end{figure}

We kept the four ROIs with highest scores (see 
figure~\ref{fig:scores_brain}). These are: Anterior Superior Temporal
Sulcus (aSTS), Temporal Pole (TP), Inferior Frontal Gyrus Orbitalis
(IFGorb) and Inferior Frontal Gyrus triangularis (IFG tri).

\begin{figure}[!t]
\centering
\includegraphics[width=2.5in]{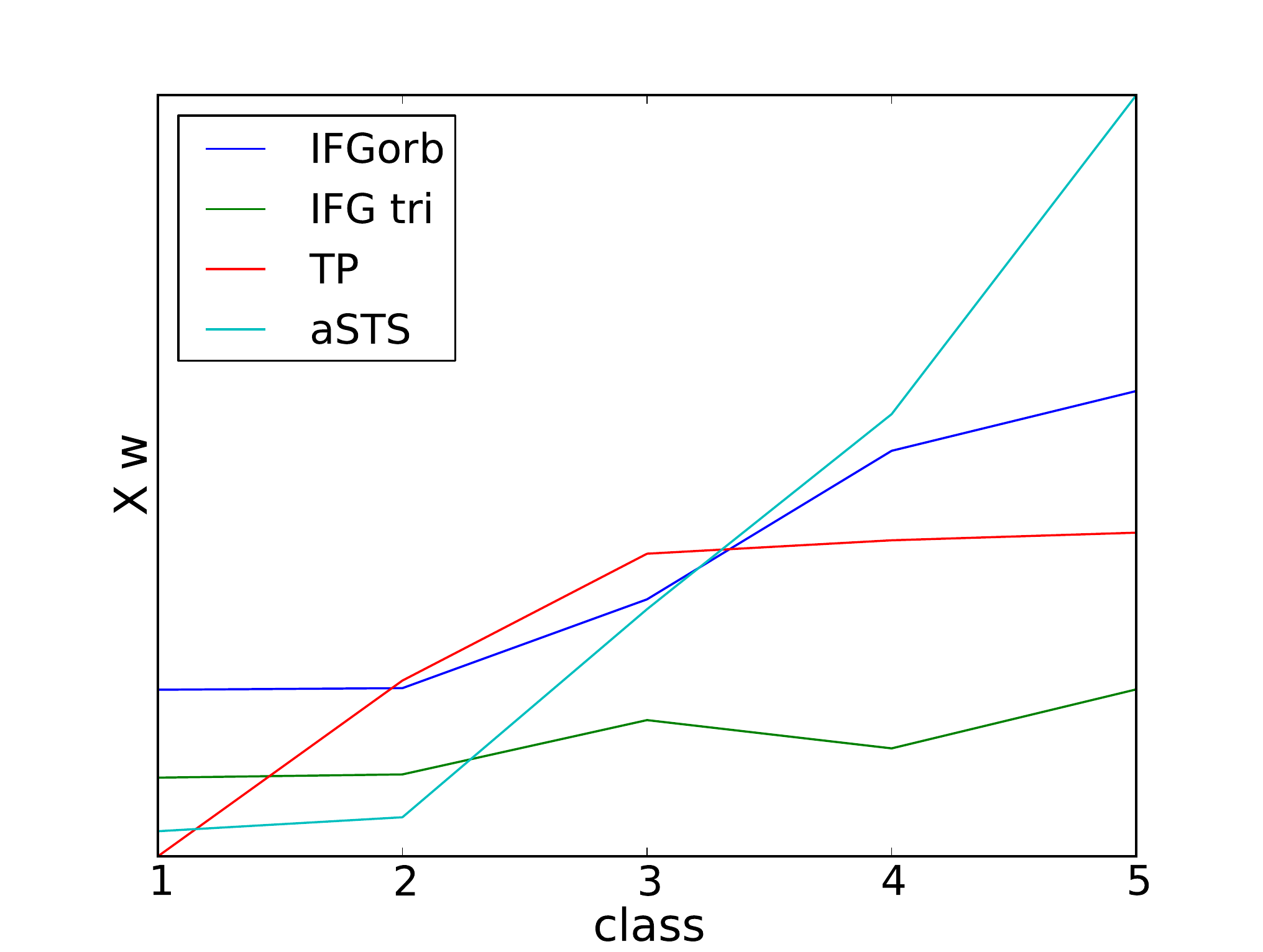}
        \caption{Data projected along $\B{\hat{w}}$ showing the
          non-linear effect in the 4 regions of interest. This
          projection gives an insight on the relationship between the
          BOLD signal and the explained variable. We observe that the
          the shape of the non-linearity varies across brain
          regions. Apart from IFG tri, all regions show a saturation
          effect in the BOLD response.}
\label{fig:bold_relationship}
\end{figure}

In order to inspect the properties of the estimated functions $F$
for each ROI, we estimated $\hat{\B{w}}$ using a pairwise logistic model.
We then projected our data $\B{X}$ along this vector $\hat{\B{w}}$ and
regularized the result using non parametric locally weighted scatterplot smoothing
(LOWESS). Results in figure~\ref{fig:bold_relationship} show that the
linearity varies in shape across ROIs which suggests that different
regions exhibit different sensitivities to the complexity parameter
under investigation. We see however a trend in the figures towards
non-linear and non-decreasing functions with some saturation
effect of the BOLD signal as in the temporal pole (TP).


In the case of the Temporal Pole (TP), which is the ROI revealing the
highest saturation effect, an F-test on the data $(\B{X} \hat{\B{w}},
y)$ reveals that the quadratic polynomial model fits better the data
than a linear model (p-value $< 0.03$). As shown in the simulations,
in this particular case pairwise loss functions are likely to improve
the recovery of active brain regions.

\section{Conclusion}

In this paper, we investigated the use of pairwise loss functions
to improve the problem of brain pattern recovery with supervised
learning applied to fMRI data. Through
simulations, we showed the benefit of such loss functions
when the target to predict is non-linearly related to the voxel
amplitudes. Experimental results on fMRI data confirmed the presence
of such non-linear effects in the data which suggest that the pairwise
approach should improve the identification of predictive brain patterns
on experimental data.

This work shows that improvements in recovery of brain activation
patterns should not only rely on the choice of a particular
regularizer, but also on an appropriate loss function.
Here we have only considered $\ell_2$-penalized models, but a natural
extension to work with full brain data
would be to consider pairwise loss functions combined with
sparse structured penalizations which incorporate domain-specific
knowledge.
This opens the path to further improvements and
refinements in the recovery of brain pattern activation via supervised
learning.

\section*{Acknowledgment}

This work was supported by the ViMAGINE ANR-08-BLAN-0250-02,
IRMGroup ANR-10-BLAN-0126-02 and Construct ANR grants.


\bibliographystyle{IEEEtran}
%

\bibliography{biblio}{}

\end{document}